\documentclass[final]{cvpr}

\usepackage{times}
\usepackage{epsfig}
\usepackage{graphicx}
\usepackage{amsmath}
\usepackage{amssymb}
\usepackage{xcolor}
\usepackage{subcaption}

\def\H{\mathbf{H}}
\def\I{\mathbf{I}}

\def\P{\mathbf{P}}

\def\U{\mathbf{U}}
\def\V{\mathbf{V}}

\def\e{\mathbf{e}}
\def\f{\mathbf{f}}
\def\g{\mathbf{g}}

\def\u{\mathbf{u}}

\def\x{\mathbf{x}}
\def\y{\mathbf{y}}
\def\z{\mathbf{z}}

\def\bSigma{\boldsymbol{\Sigma}}

\def\btheta{\boldsymbol{\theta}}

\def\diag{\textrm{diag}}

\def\argmin#1{\underset{#1}{\textrm{argmin}}}

\def\minim#1{\underset{#1}{\textrm{min}}}
\def\Exp{\mathbb{E}}

\newif\ifdraft
\drafttrue

\ifdraft
\newcommand{\rgc}[1]{{\color{red}[\textbf{RG:} #1]}}
\newcommand{\sahc}[1]{{\color{orange}[\textbf{SAH:} #1]}}
\newcommand{\ttc}[1]{{\color{blue}[\textbf{TT:} #1]}}

\else
\newcommand{\rgc}[1]{}
\newcommand{\sahc}[1]{}
\newcommand{\ttc}[1]{}

\fi

\usepackage[pagebackref=true,breaklinks=true,colorlinks,bookmarks=false]{hyperref}

\def\cvprPaperID{10065} 
\def\confYear{CVPR 2021}

\begin{document}

\title{Image Restoration by Deep Projected GSURE}

\author{Shady Abu-Hussein\\
Tel-Aviv University\\
{\tt\small shadya@mail.tau.ac.il}
\and
Tom Tirer\\
Tel-Aviv University\\
{\tt\small tomtirer@mail.tau.ac.il}
\and
Se Young Chun\\
Ulsan National Institute of \\Science and Technology\\
{\tt\small sychun@unist.ac.kr}
\and
Yonina C. Eldar\\
Weizmann Institute of Science\\
{\tt\small  yonina.eldar@weizmann.ac.il}
\and
Raja Giryes\\
Tel-Aviv University\\
{\tt\small  raja@tauex.tau.ac.il}
}

\maketitle

\begin{abstract}
Ill-posed inverse problems appear in many image processing applications, such as deblurring and super-resolution.
In recent years, solutions that are based on deep Convolutional Neural Networks (CNNs) have shown great promise.
Yet, most of these techniques, which train CNNs using external data, are restricted to the observation models that have been used in the training phase.
A recent alternative that does not have this drawback relies on learning the target image using internal learning. One such prominent example is the Deep Image Prior (DIP) technique that trains a network directly on the input image with a least-squares loss. 
In this paper, we propose a new image restoration framework that is based on minimizing a loss function that includes a "projected-version" of the Generalized Stein Unbiased Risk Estimator (GSURE) and parameterization of the latent image by a CNN. 
We demonstrate two ways to use our framework. In the first one, where no explicit prior is used, we show that the proposed approach outperforms other internal learning
methods, such as DIP. 
In the second one, we show that our GSURE-based loss leads to improved performance when used within a plug-and-play priors scheme.
\end{abstract}

\section{Introduction}
Inverse problems appear in many image processing applications, where the reconstruction of an unknown latent image $\x\in \mathbb{R}^n$ from its given corrupted version $\y\in \mathbb{R}^m$ is required. 
In many image-restoration tasks the observed image $\y$ can be expressed by the following linear model
\begin{equation}
\label{eq:obsrv_model}
    \y = \H \x + \e,
\end{equation}
where $\H \in \mathbb{R}^{m\times n}$ is the measurement operator with $m\leq n$, and $\e\sim\mathcal{N}(0, \sigma^2\I_m)$ is an additive white Gaussian noise. 
\begin{figure}[ht!]
    \centering
    \begin{subfigure}{0.45\linewidth}
        \centering
        \includegraphics[trim={50pt, 280pt, 300pt, 100pt},clip,width=100pt]{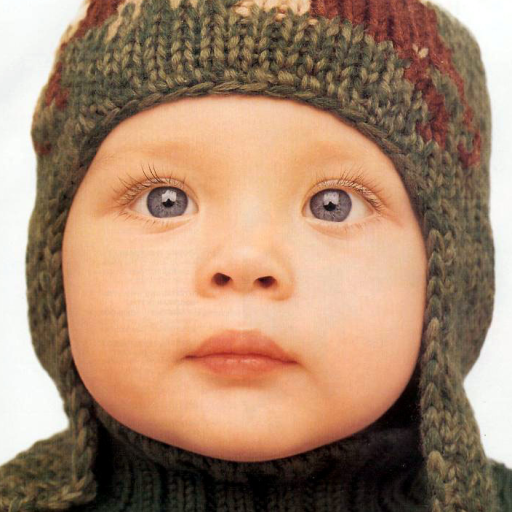}
        \caption{Original image (cropped)}
    \end{subfigure}
    \begin{subfigure}{0.45\linewidth}
        \centering
        \includegraphics[trim={50pt, 280pt, 300pt, 100pt}, clip,width=100pt]{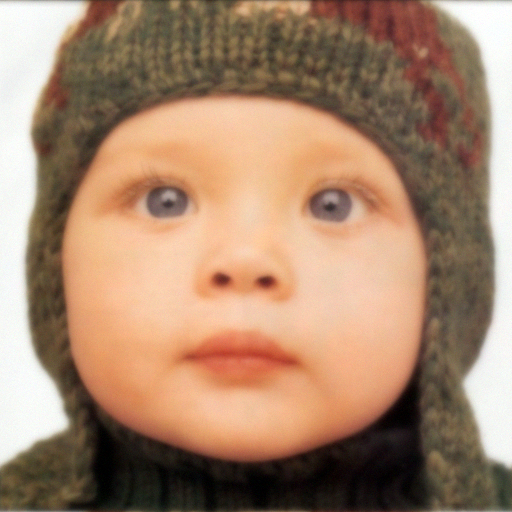}
        \caption{Blurred}
    \end{subfigure}
    \begin{subfigure}{0.45\linewidth}
        \centering
        \includegraphics[trim={50pt, 280pt, 300pt, 100pt} ,clip,width=100pt]{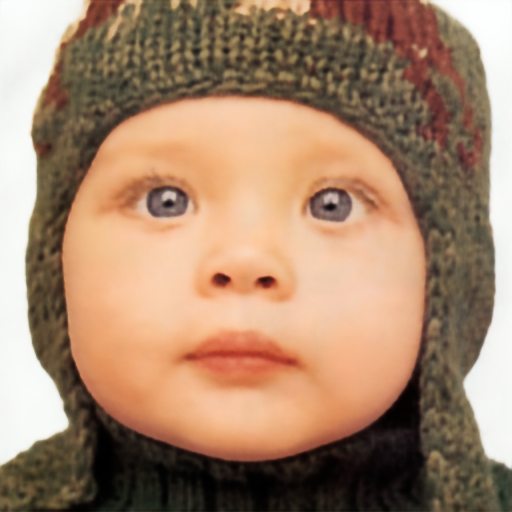}
        \caption{DIP}   
    \end{subfigure}
    \begin{subfigure}{0.45\linewidth}
        \centering
        \includegraphics[trim={50pt, 280pt, 300pt, 100pt} ,clip,width=100pt]{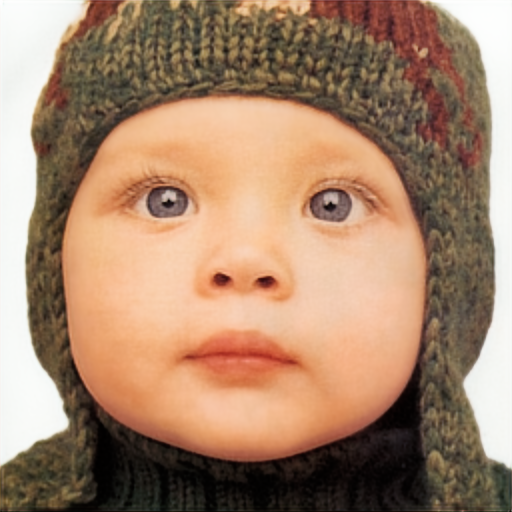}
        \caption{P-GSURE}       
    \end{subfigure}
    \begin{subfigure}{0.45\linewidth}
        \centering
        \includegraphics[trim={50pt, 280pt, 300pt, 100pt} ,clip,width=100pt]{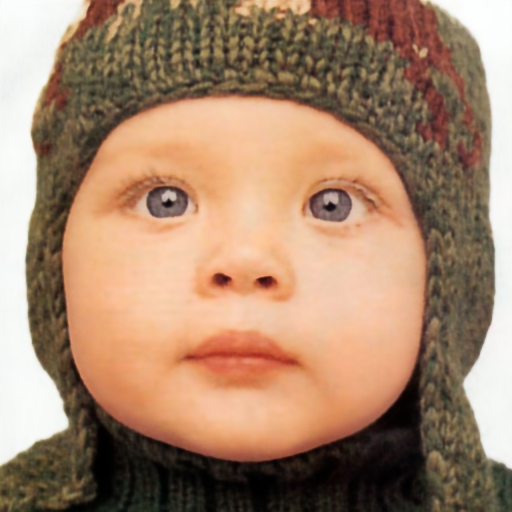}
        \caption{DIP-P\&P}
        \label{ttttt}         
    \end{subfigure}
    \begin{subfigure}{0.45\linewidth}
        \centering
        \includegraphics[trim={50pt, 280pt, 300pt, 100pt} ,clip,width=100pt]{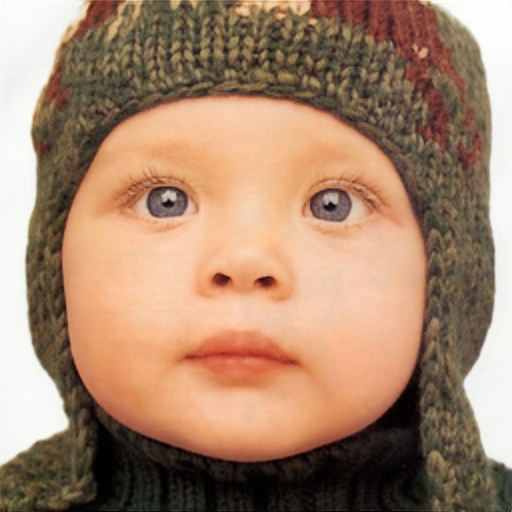}
        \caption{P-GSURE P\&P}
    \end{subfigure}
    \caption{Deblurring of the baby image from Set5, blurred using the scenario 2 model from Table \ref{table:deblurScenarios}. (a) The original image (cropped) (b) The observed blurry and noisy image (c) Reconstruction using DIP (d) Reconstruction using P-GSURE (proposed method) (e) Reconstruction using DIP based  P\&P-ADMM (with BM3D prior) (f) Reconstruction using P-GSURE based P\&P-ADMM (proposed method).}
    \label{fig:comparison_deblur_scenario2}
\end{figure}\\
For example, when $\H$ is a blur operator we refer to restoring $\x$ from $\y$ as a deblurring problem, and when $\H$ is an anti-aliasing filtering followed by sub-sampling we refer to it as super-resolution (SR).

Image restoration problems represented by \eqref{eq:obsrv_model} are usually ill-posed, in the sense that fitting the measurements $\y$ alone is not enough for a successful reconstruction of $\x$. 
Therefore, the use of some prior assumptions on $\x$ is inevitable. 
Accordingly, many reconstruction approaches are based on a minimization of a cost function formulated as
\begin{equation}
    \label{eq:reg_obj}
    \ell(\tilde{\x}; \y) + \beta s(\tilde{\x}),
\end{equation}
where 
$\tilde{\x}$ is the optimization variable, 
$\ell(\cdot;\y)$ is the fidelity term, $s(\cdot)$ is the prior / regularization term, and $\beta > 0$ controls the strength of the regularization. The fidelity term assures compliance to the measurements $\y$, while the prior term encourages the reconstruction to satisfy some prior assumptions on $\x$, such as sparsity or non-local repeating patterns.

Over the past decade, Convolutional Neural Networks (CNNs) have shown great potential when it comes to solving imaging inverse problems 
as they circumvent the hardness of designing priors for natural images  \cite{lim2017enhanced,dong2015image, Sun2015LearningAC,Nah2016DeepMC}.
However, in order to perform well, the CNNs require that the observation model at test time is the exact model used in the training phase. For instance, in the SR task, it is common to train the CNNs to reconstruct $\x$ from $\y$ that is observed using the bicubic down-sampling kernel \cite{dong2015image,lim2017enhanced}. But when examined on other down-sampling kernels these networks perform very poorly 
\cite{shocher2018zero,tirer2019super,Hussein_corr_flt}.

An alternative to the offline training of CNNs with a pre-determined observation model (which may not match the model at test time) is ``internal learning", \ie training that is based only on the given observed image \cite{shocher2018zero,ulyanov2018deep}.
One popular internal learning method is Deep Image Prior (DIP) \cite{ulyanov2018deep}, where a CNN with a set of parameters $\theta$ (and a fixed random input) is employed to parameterize the latent image $\tilde{\x}=\f(\btheta)$. It is then used to minimize the Least-Squares (LS) loss (with no use of an explicit prior $s(\cdot)$), $$\ell_{\text{\scriptsize{LS}}}(\f(\btheta))=\|\y-\H \f(\btheta) \|_2^2.$$ 

Using such a scheme implicitly benefits from the fact that the CNN architecture
promotes non-local repeating patterns, which are common in natural images.
However, because DIP uses an over-parameterized CNN
it may overfit the noise.
Therefore, it requires an accurate early stopping, and 
its performance degrades dramatically when faced with heavy noise.
This 
inspired follow-up works to incorporate DIP with additional explicit regularizations \cite{Cheng2019Bayesian, liu2019image, mataev2019deepred, zukerman2020bp}.

In this paper, we propose a new image restoration framework that is based on minimizing a loss function, where instead of a plain fidelity term $\ell(\tilde{\x})$, we use a "projected-version" of the Generalized Stein Unbiased Risk Estimator (GSURE) \cite{Stein:1981vf,Eldar2009fga} together with a parameterization of the latent image by a CNN. 
We demonstrate two ways to use our framework. In the first one, where no explicit prior is used, we show that the proposed approach outperforms other internal learning
methods, such as DIP \cite{ulyanov2018deep}
and ZSSR \cite{shocher2018zero}. 
In the second one, we show that our GSURE-based loss leads to improved performance when used within a plug-and-play priors scheme \cite{venkatakrishnan2013plug, RED, PnP_ADMM,tirer2018image,mataev2019deepred}, where an additional prior $s(\cdot)$ is imposed through the application of existing denoisers.

\section{Related work}

The self-similarity of natural images (\ie the phenomenon of recurrence of patterns within and across scales) has been shown to be a powerful prior for image reconstruction \cite{buades2005non,meyer2001oscillating,dabov2007image,glasner2009super, huang2015single}.
Recent empirical works, such as DIP \cite{ulyanov2018deep} and zero-shot super-resolution (ZSSR) \cite{shocher2018zero} showed that the self-similarity prior imposed by CNN architectures is sufficient for recovering an image simply by training a deep neural network from scratch at test time using only the observed degraded image.
While ZSSR focused only on SR, the approach in DIP considered general inverse problems and led to many follow-up papers \cite{DoubleDIP, Zhou2018, shaham2019singan, InGAN, sun2019test,williams2019deep,Hanocka20Point2Mesh}.

In the presence of noise, DIP requires an accurate early stopping to avoid overfitting the noise. To mitigate  this issue, and further improve the performance, several papers incorporate it with additional explicit priors \cite{Cheng2019Bayesian, liu2019image, mataev2019deepred, zukerman2020bp}.

Another line of work, which is related to our paper, focuses on offline unsupervised learning: training CNNs on noisy training sets (without ground truth clean images).
One such work is 
Noise2Noise \cite{noise2noise}, where the training set includes two independent noise realizations per image,
and the strategy is examined 
for various noise models and some applications, including compressive sensing MR recovery~\cite{noise2noise}.
Another method, proposed in \cite{soltanayev2018training}, is based on Stein’s unbiased risk estimator (SURE) \cite{Stein:1981vf} (which is an unbiased estimate for the estimation MSE when the ground truth is unknown).
This method trains CNNs with a training set of a single noise realization per image and SURE as the loss function. Yet, similarly to the original SURE \cite{Stein:1981vf}, it is limited to Gaussian denoising.
Another paper that uses SURE to train CNNs for Gaussian denoising is \cite{Metzler:2019gsure}, where the performance is also demonstrated when training  using only a single noisy image. These works have been extended to train networks (without ground truth data) for image recovery from undersampled compressive sensing measurements but the training is performed in this case using multiple images.

Over the years, SURE has been extended to more general noise models such as mixed Poisson-Gaussian model~\cite{LeMontagner:2014jd} or a general exponential family~\cite{Eldar2009fga}. The Generalized SURE (GSURE) in \cite{Eldar2009fga}
has been used for offline unsupervised training (without ground truth data) of networks for image recovery from undersampled compressive sensing measurements \cite{Metzler:2019gsure}. Another work, combined the regular SURE with the LDAMP technique \cite{Metzler17LDAMP} for solving the same problem using also offline training \cite{Zhussip_2019_CVPR}.
In contrast to these two approaches that perform offline training on a large set of images with pre-determined observation models, in this paper we will use GSURE to solve ill-posed problems by training a CNN using only the observed image.

Using SURE to train a CNN requires evaluating the divergence of the network's output with respect to its input, as well as the gradient of the divergence with respect to the network's parameters.
Most previous works~\cite{soltanayev2018training,Zhussip_2019_CVPR,Metzler:2019gsure} have used Monte-Carlo (MC) approximation of the divergence, as proposed in \cite{Ramani2008Monte}, due to fast and easy computation, which we employ also in our paper. Alternative approximation of the divergence, based on random projections of the network's output, has been shown in \cite{Soltanayev2020Divergence}.

The image restoration framework, presented in this paper, is demonstrated with no explicit prior (similarly to DIP), and also when an additional prior is used.
In the latter case, we are inspired by the Plug-and-Play Priors (P\&P) approach \cite{venkatakrishnan2013plug,metzler2016denoising, RED, PnP_ADMM,kamilov2017plug,tirer2018image}, where the prior used for solving different inverse problems is imposed through the application of denoising operations. Specifically, in this paper, we impose the additional prior using the BM3D denoiser \cite{dabov2007image}. Note that several P\&P methods have also shown excellent results using offline trained CNN denoisers \cite{Meinhardt17Learning,tirer2018image,zhang2017learning,tirer2018image}.
We note that there exists a method \cite{mataev2019deepred} that combines DIP and P\&P.
The major difference between our approach and \cite{mataev2019deepred} is in the fact that \cite{mataev2019deepred} uses the plain LS fidelity term while we use a GSURE-based loss function. GSURE has the advantage over DIP that it leads to a more stable training that is less sensitive than DIP to the parameters of the network and the early stopping in the training.

\section{Method}

In this paper, we employ the generalized SURE (GSURE), which has been developed in \cite{Eldar2009fga}.
Let $\hat{\x}(\y)$ be a (weakly) differentiable estimator of the deterministic unknown $\x$.
The original SURE formula \cite{Stein:1981vf} provides an unbiased estimate of the MSE, $\Exp{\|\hat{\x}-\x\|_2^2}$, for the case where $\e$ is white Gaussian noise and $\H$ equals the identity matrix $\I_n$.
The work in \cite{Eldar2009fga} has proposed the project GSURE, which generalizes the MSE estimate to more cases, where one of them is the ill-posed linear model that is considered in Equation~\eqref{eq:obsrv_model}.

While previous works used this ``projected" GSURE mainly for tuning only a few hyper-parameters of a pre-determined estimator $\hat{\x}$ (\ie by setting them to values that minimize Equation~\eqref{eq:gsure}) \cite{Giryes11Projected}, in this paper, we proposed to  extend the role that GSURE plays in estimating $\x$.
Specifically, we propose to estimate $\x$ (from scratch) as the minimizer of the cost function
\begin{equation}
\label{eq:gsure_cost}
    \ell_{\text{\scriptsize{GSURE}}}(\tilde{\x}) + \beta s(\tilde{\x}),
\end{equation}
where 
$\tilde{\x}$ is the optimization variable, 
$s(\tilde{\x})$ is a prior term that can improve the estimation and $\beta$ is a positive hyper-parameter that balances the two terms.

Before introducing our method, we first discuss the use of the projected GSURE and its formulation. Then we show how it can be used for network training from a single example. This is followed by a derivation of a plug and play based approach allows incorporating prior knowledge through a denoiser to the single image training. Finally, we draw a connection between our formulation and other recently used loss functions.

\subsection{The projected GSURE}

In the case where $m \leq n$ and $\H$ has a full row rank, its pseudoinverse is given by $\H^\dagger:=\H^T(\H\H^T)^{-1}$ and the projection onto $\mathcal{R}(\H^T)$ (\ie the subspace spanned by the rows of $\H$) is given by $\P_H:=\H^\dagger\H$. As the observation model is ill-posed, the GSURE provides an estimate of the MSE only in $\mathcal{R}(\H^T)$, i.e., of the projected MSE $\Exp{\|\P_H\hat{\x}-\P_H\x\|_2^2}$. The GSURE estimate is given by
\begin{align}
\label{eq:gsure}
    \ell_{\text{\scriptsize{GSURE}}}(\hat{\x}(\u)) =& c + \|\P_H \hat{\x}(\u)\|_2^2 - 2\hat{\x}^T(\u)\hat{\x}_{\text{\scriptsize{ML}}} \\ \nonumber
    &+ 2 \mathrm{div}_\u (\P_H \hat{\x}(\u)),
\end{align}
where $c$ is a term that does not depend on $\hat{\x}$, 
$\u \in \mathbb{R}^n$ is a sufficient statistic (for the estimation of $\x$ from $\y$) that $\hat{\x}$ gets as input, $\hat{\x}_{\text{\scriptsize{ML}}}$ is the maximum likelihood estimator and the last term denotes the divergence
\begin{equation}
\label{eq:div}
    \mathrm{div}_\u (\P_H \hat{\x}(\u)) = \sum \limits_{i=1}^{n} \frac{\partial [\P_H \hat{\x}(\u)]_i}{\partial u_i}.
\end{equation}
For the considered model, typical choices are $\u=\frac{1}{\sigma^2}\H^T\y$\footnote{Notice that although the GSURE relies on knowing the noise level $\sigma$, $\sigma$ can be estimated from the $\y$ using techniques such as \cite{liu2006noise}.} and $\hat{\x}_{\text{\scriptsize{ML}}} = \H^\dagger\y$.

In practice, the assumption that $\H$ has a full row-rank does not hold always. Moreover, some of the singular values of $\H$ may be very close to zero such that even if $\H$ has a full row-rank, numerically the inversion is unstable and therefore leads to poor MSE estimations. This is particularly important in the deblurring problem where many of the blur operators have a full rank in theory but in practice, many of their singular values are close to zero and therefore should be treated as such. In this case, to calculate the pseudo-inverse of $\H$, we take the SVD decomposition $\H=\U\bSigma \V^T$, where $\bSigma = \diag(\sigma_1\, \dots, \sigma_m, 0, \dots, 0)$, and set the pseudo inverse to be $\H^\dag=\V\bSigma^\dag \U^T$. The pseudo inverse $\bSigma^\dag$, which is a diagonal matrix, is defined such that $(\bSigma^\dag)_{i,i} = 1/\sigma_i$ if $i\le m$ and $\sigma_i > \xi$ (for a given small threshold $\xi$) and $(\bSigma^\dag)_{i,i} = 0$ otherwise. 

This formulation for the Pseudo-inverse can be easily computed in the deblurring case (with circular padding) since the blur kernels are shift-invariant and therefore they are diagonalized by the Fourier transform. Thus, to apply the Pseudo-inverse of $\H$ we can simply apply the Fourier transform on the input, then multiply it with the inverse eigenvalues of the kernel (that are larger than the threshold $\xi$) and then apply the inverse Fourier transform.

Notice that the formula of the projected GSURE remains the same also when we use the thresholding of the small singular values. The only difference is in the calculation of $\H^\dag$, which then sets also $\P_H$ in Equation~\eqref{eq:gsure}.

Another element in Equation~\eqref{eq:gsure} that deserves a special care is the divergence term. It is essentially the only term in Equation~\eqref{eq:gsure} that requires connecting $\tilde{\x}$ with an input $\u$. 
In this paper we resolve this issue by parameterizing $\tilde{\x}$ by a CNN with weights $\btheta$ and input $\u=\frac{1}{\sigma^2}\H^T\y$, \ie $\tilde{\x}=\f(\u;\btheta)$.
While this differentiable connection between $\tilde{\x}$ and $\u$ allows to compute the derivatives required in Equation~\eqref{eq:div}, 
to ease the computation we use the MC approximation of \cite{Ramani2008Monte}:
\begin{equation}
\label{eq:div_mc}
    \mathrm{div}_\u (\P_H \tilde{\x}(\u)) \approx \g^T \P_H \frac{\tilde{\x}(\u + \epsilon \g) - \tilde{\x}(\u)}{\epsilon},
\end{equation}
where $\g$ is a Gaussian vector drawn from $\mathcal{N}(\mathbf{0},\I_n)$ and $\epsilon$ is a fixed small positive value.

To summarize, our framework for estimating $\x$ is based on optimizing the weights of the CNN $\f(\u;\btheta)$ by minimizing the loss function
\begin{align}
\label{eq:gsure_cost_cnn}
    L(\btheta) = \tilde{\ell}_{\text{\scriptsize{GSURE}}}(\f(\u;\btheta)) + \beta s(\f(\u;\btheta)),
\end{align}
where    
\begin{align}
\label{eq:gsure_cost_cnn2}    
    \tilde{\ell}_{\text{\scriptsize{GSURE}}}(\f(\u;\btheta))=& \|\P_H \f(\u;\btheta) \|_2^2 - 2\f^T(\u;\btheta) \H^\dagger \y \\ \nonumber
    &+ 2\g^T \P_H \frac{\f(\u + \epsilon \g;\btheta) - \f(\u;\btheta)}{\epsilon}.
\end{align}
As mentioned above, in tasks like deblurring and super-resolution, the operators $\H, \H^\dagger$ and $\P_H$ have a fast implementation using the Fast Fourier Transform (FFT).
In addition, fast automatic differentiation of $\tilde{\ell}_{\text{\scriptsize{GSURE}}}(\f(\u;\btheta))$ with respect to $\btheta$ can be obtained using popular software packages, such as TensorFlow and PyTorch. Note that the random variable $\g$ is drawn in each iterations of the optimization (this improves the expectation of the MC approximation).
After obtaining the optimized weights, $\hat{\btheta}$, the latent image is estimated by $\hat{\x}=\f(\u;\hat{\btheta})$.

We now turn to describe two instantiations of Equation~\eqref{eq:gsure_cost_cnn}: one without the prior function $s(\cdot)$ and the second when the additional prior is implicitly imposed by a ``plug-and-play" denoiser.

\subsection{GSURE-based network training}

Standard neural network training techniques for inverse problems in image processing require a large amount of training data as they are usually trained based on pairs of the input deteriorated image $\y$ and the true clean image $\x$. Under the assumption that the image statistics in the training resemble the one of the input image to the network, then the fact that the network was trained to minimize the MSE on the training images is expected to lead to a small MSE also on the input test image. The GSURE gives us an opportunity to treat the MSE of the input image directly as it provides an unbiased estimate for it. This is particularly important in cases where the image statistics or the measurement model are different than the training setup. 

Having the estimation of the GSURE for the MSE (or projected MSE), we use it to train a network just a single image by directly minimizing its (estimated projected) MSE. Notice that although we do not use any explicit image prior in this setting, the fact that we use a neural network for the reconstruction implicitly imposes a prior as shown in the DIP approach \cite{ulyanov2018deep}.
Moreover, the implicit regularization of optimization methods like SGD and Adam is another source of regularization \cite{Soltanolkotabi19Theoretical}. 
Therefore, the first reconstruction method that we examine in this paper is solely based on minimizing $L(\btheta) = \tilde{\ell}_{\text{\scriptsize{GSURE}}}(\f(\u;\btheta))$ using Adam \cite{kingma2014adam}.

While the GSURE based approach bears similarity to the original DIP work, it differs from it in several ways. First, it is theoretically motivated by minimizing an estimate of the (projected) MSE. Second, it uses a different loss function --- $\tilde{\ell}_{\text{\scriptsize{GSURE}}}(\f(\u;\btheta))$ instead of $\ell_{\text{\scriptsize{LS}}}(\f(\u;\btheta))$. Third, it uses a different input for the CNN --- the sufficient statistic $\u$ rather than a large tensor of random noise. As will be shown in the experiments section, GSURE based training not only outperforms the original DIP in terms of PSNR but also does not require accurate early stopping to avoid noise overfitting.
Another related work is a recent paper \cite{zukerman2020bp} that solves deblurring tasks using DIP with the BP loss (instead of the plain LS loss as in regular DIP), 
which is equivalent to omitting the divergence term in GSURE (this also allows their CNN to get random noise as input).  Yet, the divergence term that we have in the GSURE stabilizes the effect of the noise on the BP loss, which as shown in \cite{zukerman2020bp} is sensitive to noise and requires accurate early stopping.
We discuss more the relationship to this loss in Section~\ref{sec:other_loss}.

\subsection{GSURE-based P\&P-ADMM}

In an ideal case, where GSURE would provide a perfect estimation for the MSE, it would have been sufficient to train a neural network using GSURE to get ``perfect'' results. Yet, it is only an estimate of the MSE and with the projection it estimates only of the projected MSE. Thus, using it alone for minimization is not sufficient and adding another image prior $s(\cdot)$ can lead to improved reconstruction.

To achieve this goal, we follow the P\&P denoisers concept \cite{venkatakrishnan2013plug}.
We modify Equation~\eqref{eq:gsure_cost_cnn} using variable splitting and obtain the following constrained optimization problem
\begin{align}
\label{eq:gsure_cost_cnn_constr}
    \minim{\btheta,\z} \,\,\,\, & \tilde{\ell}_{\text{\scriptsize{GSURE}}}(\f(\u;\btheta)) + \beta s(\z) \\ \nonumber
    & \mathrm{s.t.} \,\,\,\, \z = \f(\u;\btheta).
\end{align}
This problem can be solved using ADMM (a review on ADMM can be found in \cite{boyd2011distributed}).
Specifically, we construct the augmented Lagrangian 
\begin{align}
\label{Eq_pnp_AL}
L_\rho =& \,\, \tilde{\ell}_{\text{\scriptsize{GSURE}}}(\f(\u;\btheta)) + \beta s(\z) + \tilde{\mathbf{v}}^T(\f(\u;\btheta)-\z) \nonumber \\
&+ \frac{\rho}{2} \| \f(\u;\btheta)-\z \|_2^2 \nonumber \\
=&  \,\, \tilde{\ell}_{\text{\scriptsize{GSURE}}}(\f(\u;\btheta)) + \beta s(\z) + \frac{\rho}{2} \| \f(\u;\btheta)-\z+\mathbf{v}\|_2^2 \nonumber \\
&- \frac{\rho}{2} \|  \mathbf{v} \|_2^2,
\end{align}
where $\tilde{\mathbf{v}}$ is the dual variable, $\mathbf{v} = \frac{1}{\rho} \tilde{\mathbf{v}}$ is the scaled dual variable, and $\rho$ is the ADMM penalty parameter. 
Then, we initialize $\mathbf{v}_0=\mathbf{0}$ and
$\z_0=\hat{\x}_{\text{\scriptsize{ML}}}$, and repeat the following three steps for $k=1,\ldots,N_{iter}$
\begin{align}
\label{eq:gsure_admm}
\btheta_k &= \argmin{\btheta} \,\,\, \tilde{\ell}_{\text{\scriptsize{GSURE}}}(\f(\u;\btheta)) \nonumber \\ & ~~~~~~~~~~~~~~~~~~+ \frac{\rho}{2} \| \f(\u;\btheta) - \z_{k-1} + \mathbf{v}_{k-1} \|_2^2,  \nonumber \\
\z_k &= \argmin{\z} \,\,\, \frac{1}{2\beta/\rho} \| \f(\u;\btheta_k) + \mathbf{v}_{k-1} - \z \|_2^2 + s(\z),  \nonumber \\
\mathbf{v}_k &= \mathbf{v}_{k-1} + \f(\u;\btheta_k) - \z_k.
\end{align}
The first step can be optimized via Adam with warm-start initialization (given by the previous $\btheta_{k-1}$). The second step describes obtaining $\z_k$ using a denoiser
for white Gaussian noise of variance $\beta/\rho$ applied on $\f(\u;\btheta_k) + \mathbf{v}_{k-1}$. 
Following the P\&P denoisers concept, in our experiments we implement this step by an existing denoiser $\z_k = \mathcal{D}(\f(\u;\btheta_k) + \mathbf{v}_{k-1}; \sqrt{\beta/\rho})$ (e.g., colored BM3D \cite{dabov2007image}) that implicitly imposes the prior $s(\cdot)$ on the reconstruction.

Note that in typical P\&P schemes, which use plain fidelity terms, the first step in Equation~\eqref{eq:gsure_admm} inverts the observation model with very minor regularization (the simple $\ell_2$-norm term that is weighted by $\rho/2$). 
In contrast, in our approach, this step enjoys both the advantage of GSURE over LS (\eg better handling the noise via the divergence term) and the implicit prior of the CNN architecture.

\subsection{Relation to other loss functions}
\label{sec:other_loss}

Let us present a motivation for optimizing Equation~\eqref{eq:gsure_cost} (or its CNN parameterized form Equation~\eqref{eq:gsure_cost_cnn}) by comparing it with alternative cost functions.
Perhaps the most common approach to estimate $\x$ from observations modeled by Equation~\eqref{eq:obsrv_model} is to minimize cost function of the form $\ell_{\text{\scriptsize{LS}}}(\tilde{\x}) + \beta s(\tilde{\x})$, where $\ell_{\text{\scriptsize{LS}}}(\tilde{\x})=\|\y-\H\tilde{\x}\|_2^2$ is the least squares (LS) fidelity term \cite{mataev2019deepred,venkatakrishnan2013plug,zhang2017learning}.

Recently, an alternative fidelity term $\ell_{\text{\scriptsize{BP}}}(\tilde{\x})=\|\H^\dagger(\y-\H\tilde{\x})\|_2^2$, dubbed ``back-projection" (BP) term, has been shown to have advantages over LS for different ill-posed problems, such as deblurring and super-resolution, where the condition number of $\H$, \ie the ratio between the largest and smallest singular values of $\H$, is large \cite{tirer2018image,tirer2020back,tirer2020convergence}. However, the tradeoff is that the BP term is more sensitive than LS to noise amplification when $\H$ has  small singular values. 

Now, observe that if we ignore the 
divergence term in Equation~\eqref{eq:gsure} then minimizing $\ell_{\text{\scriptsize{GSURE}}}(\tilde{\x})$ (where $\hat{\x}_{\text{\scriptsize{ML}}} = \H^\dagger\y$) is equivalent to minimizing $\ell_{\text{\scriptsize{BP}}}(\tilde{\x})$. This follows from (recall that $\P_H=\P_H^T$ and $\P_H \H^\dagger = \H^\dagger$) the fact that
\begin{align}
\label{eq:bp_gsure_connection}
    &\argmin{\tilde{\x}} \,\,\, \|\H^\dagger(\y-\H\tilde{\x})\|_2^2 \\ \nonumber
    &= \argmin{\tilde{\x}} \,\,\, \|\P_H\tilde{\x}\|_2^2 - 2 \tilde{\x}^T \P_H \H^\dagger\y + \| \H^\dagger\y \|_2^2 \\ \nonumber
    &= \argmin{\tilde{\x}} \,\,\, \|\P_H\tilde{\x}\|_2^2 - 2 \tilde{\x}^T \H^\dagger\y.
\end{align}
Therefore, $\ell_{\text{\scriptsize{GSURE}}}(\tilde{\x})$ is expected to have advantages over $\ell_{\text{\scriptsize{LS}}}(\tilde{\x})$ that are similar to those of $\ell_{\text{\scriptsize{BP}}}(\tilde{\x})$. Moreover, note that the term $\mathrm{div}_\u (\P_H \tilde{\x}(\u))$ in $\ell_{\text{\scriptsize{GSURE}}}(\tilde{\x})$ constrains the changes in $\tilde{\x}$ due to perturbations in the observations, which provides a regularization against noise amplification. Therefore, $\ell_{\text{\scriptsize{GSURE}}}(\tilde{\x})$ is also expected to be more suitable than $\ell_{\text{\scriptsize{BP}}}(\tilde{\x})$ to handle high noise levels.

\begin{figure*}
    \centering
    \begin{subfigure}{0.30\linewidth}
        \centering
        \includegraphics[trim={0pt, 120pt, 120pt, 0pt},clip,width=100pt]{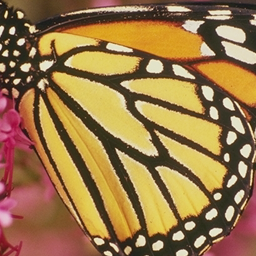}
        \caption{Original image (cropped)}      
    \end{subfigure}
    \begin{subfigure}{0.20\linewidth}
        \centering
        \includegraphics[trim={0pt, 120pt, 120pt, 0pt},clip,width=100pt]{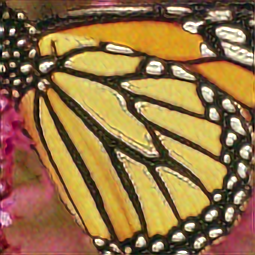}
        \caption{ZSSR}       
    \end{subfigure}
    \begin{subfigure}{0.20\linewidth}
        \centering
        \includegraphics[trim={0pt, 120pt, 120pt, 0pt},clip,width=100pt]{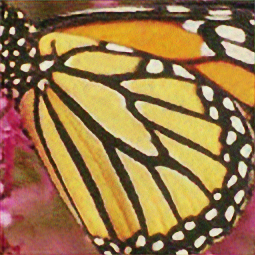}
        \caption{DIP}
    \end{subfigure}
    \begin{subfigure}{0.20\linewidth}
        \centering
        \includegraphics[trim={0pt, 120pt, 120pt, 0pt},clip,width=100pt]{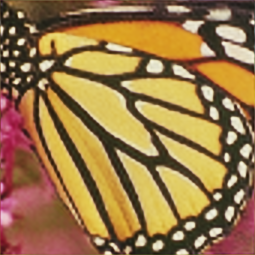}
        \caption{P-GSURE}       
    \end{subfigure}\\
    \begin{subfigure}{0.30\linewidth}
        \centering
        \includegraphics[trim={0pt, 120pt, 120pt, 0pt},clip,width=100pt]{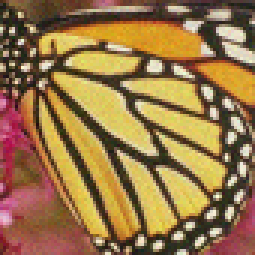}
        \caption{LR}   
    \end{subfigure}
    \begin{subfigure}{0.20\linewidth}
        \centering
        \includegraphics[trim={0pt, 120pt, 120pt, 0pt},clip,width=100pt]{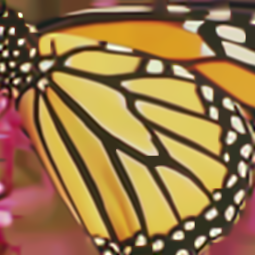}
        \caption{P\&P - BM3D}        
    \end{subfigure}
    \begin{subfigure}{0.20\linewidth}
        \centering
        \includegraphics[trim={0pt, 120pt, 120pt, 0pt},clip,width=100pt]{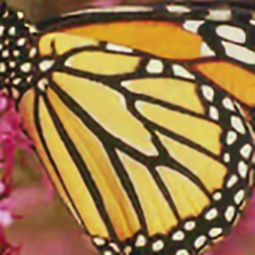}
        \caption{DIP-P\&P}        
    \end{subfigure}
    \begin{subfigure}{0.20\linewidth}
        \centering
        \includegraphics[trim={0pt, 120pt, 120pt, 0pt},clip,width=100pt]{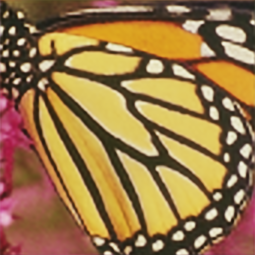}
        \caption{P-GSURE P\&P}       
    \end{subfigure}
    \caption{Super-resolution using the scenario 6 model from Table \ref{table:SRScenarios}, tested on the butterfly image from Set5, (a) The original image cropped (b) The observed low-resolution and noisy image (c) Reconstruction using ZSSR (d) Reconstruction using DIP (e) Reconstruction using P-GSURE (proposed method) (f) P\&P-ADMM method (using BM3D prior) (g) Reconstruction using DIP based P\&P-ADMM (with BM3D prior) (h) Reconstruction using P-GSURE based P\&P-ADMM (proposed method).}
    \label{fig:comparison_SR_scenario5}
\end{figure*}
\section{Experiments}
We demonstrate our method on two common image processing tasks on the Set5 and Set14 datasets. We first present the experiments for deblurring, in which a restoration of the original sharp image from the blurry and noisy observation is desired. This is followed by evaluating our approach on the SR task, where we seek to recover the original High-Resolution (HR) image from its
given Low-Resolution (LR) and noisy version. 

In each task, we examine the proposed GSURE-based method with and without an additional BM3D prior. In all the experiments, we compare our results with those obtained by using DIP instead of GSURE (recall that they mainly differ in their loss functions and in their inputs).
For both GSURE and DIP, we use the same 'skip' network used in the original DIP code published by \cite{ulyanov2018deep}. 
In the ``internal learning" SR experiments (where only the self-similarity of the observed image is used as prior), we also compare to the results of ZSSR \cite{shocher2018zero}.
In the experiments that exploit the BM3D prior, we also present the results of the plain P\&P method (that does not include the CNN parameterization) \cite{venkatakrishnan2013plug}.

\begin{figure}
    \centering
    \includegraphics[width=\linewidth]{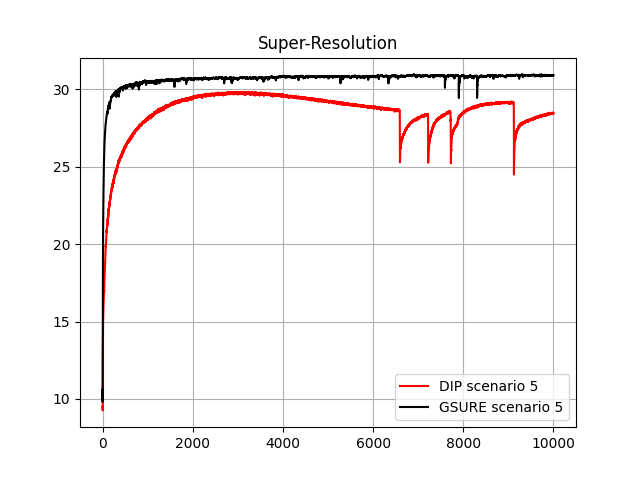}
    \caption{Super-Resolution results of scenario 5 averaged on Set5.}
    \label{fig:plot_SR}
\end{figure}

We use $\epsilon = 10^{-6}$ for approximating the divergence in \eqref{eq:div_mc} for all GSURE scenarios and optimize the network weights using ADAM \cite{kingma2014adam} with learning rate of $10^{-2}$ for both DIP and GSURE.
For achieving the best performance in DIP, we run the algorithm in all scenarios on all the images in Set5, then we average the iterations number in which the maximum PSNR is achieved, and use that as the early stopping criterion in each scenario (6 values, one for each scenario). In GSURE, however, we take the reconstruction at the minimum value of $\ell_{\scriptsize{GSURE}}(\Tilde{\x})$, since it is on par with the MSE value. Notice that unlike in DIP, with GSURE we do not need the ground truth information to set the stopping criterion, as demonstrated in Figures \ref{fig:plot_deblur} and \ref{fig:plot_SR}.
In P\&P based GSURE and DIP we take the reconstruction from the last ADMM iteration.

\begin{table*}
\small
\centering
    \caption{Deblurring results (averaged PSNR) for the different scenarios} \label{table:deblurResults}
    \begin{tabular}{| c | c | c | c | c ||  c | c | c | c | c | c |}
    \hline
                 & \multicolumn{2}{c|}{Set5} & \multicolumn{2}{c||}{Set14}   & \multicolumn{3}{c|}{Set5}    & \multicolumn{3}{c|}{Set14}         \\ \hline
    Scenario     & DIP   & P-GSURE            & DIP & P-GSURE            & P\&P             & P\&P-DIP   & P\&P-GSURE        & P\&P  & P\&P-DIP   & P\&P-GSURE\\ \hline
        1        & 30.36 & \textbf{31.95}   & 27.10  & \textbf{30.14}    & \textbf{33.36}   & 32.18 & 33.30         & 30.72 & 27.49     & \textbf{30.97}  \\ \hline
        2        & 30.24 & \textbf{31.39}   & 27.01  & \textbf{28.28}    & 31.25            & 31.5  & \textbf{31.82}& 28.27 & 27.46     & \textbf{28.92}  \\ \hline
        3        & 28.42 & \textbf{30.28}   & 25.17  & \textbf{27.75}    & \textbf{32.60}   & 29.68 & 30.38         & \textbf{30.31} & 25.85     & 27.58  \\ \hline
        4        & 31.16 & \textbf{31.99}   & 26.77  & \textbf{29.49}    & 31.85            & 31.23 & \textbf{32.53}& 29.29 & 28.85     & \textbf{29.7}  \\ \hline
        5        & 30.41 & \textbf{30.93}   & 26.91  & \textbf{28.06}    & 31.80            & 31.64 & \textbf{32.05}& 28.66 & 27.55     & \textbf{28.71}  \\ \hline
        6        & 33.44 & \textbf{35.55}   & 31.32  & \textbf{34.21}    & 34.55            & 33.32 & \textbf{35.67}& 33.69 & 29.74     & \textbf{34.45}  \\ \hline
    \end{tabular}
\end{table*}

\subsection{Deblurring}
In the image deblurring problem, $\H$ is a blur operator defined as a circular convolution with a blur filter $h$. Thus, it has fast implementation using the Fast Fourier Transform (FFT). Accordingly, $\H^T$ is a blur operator with a flipped version of the blur kernel $h$. 
As a result, calculating $\H\H^T$ can be achieved by simply multiplying the FFTs of the blur kernel and its flipped version. 

We adopt the test scenarios used in \cite{BM3D_Frames}, whose blur kernels $h$ and noise levels $\sigma$ are presented in Table~\ref{table:deblurScenarios}. Note that
in these scenarios 
the condition number of $\H$ is extremely high, hence, when computing $(\H\H^T)^\dagger$, we zero (rather than invert) all the places where the eigenvalues are smaller than $\xi$ (the specific values are given in Table \ref{table:deblurScenarios}).

\begin{table}
\small
\centering
    \caption{Deblurring test scenarios} \label{table:deblurScenarios}
    \begin{tabular}{ | c | c | c | c |}
    \hline
    Scenario     & $h(x_1, x_2)$                    & $\sigma^2$ & $\xi$ \\ \hline
        1        & $1/(1+x_1^2 + x_2^2)$            & 2          & $5\times 10^{-2}$\\ \hline
        2        & $1/(1+x_1^2 + x_2^2)$            & 8          & $1\times 10^{-1}$\\ \hline
        3        & $9 \times 9$ uniform             & 0.3        & $5\times 10^{-3}$\\ \hline
        4        & $[1,4,6,4,1]^T[1,4,6,4,1]/256$   & 49         & $1\times 10^{-1}$\\ \hline
        5        & Gaussian with $std = 1.6$        & 4          & $5\times 10^{-2}$\\ \hline
        6        & Gaussian with $std = 0.4$        & 64         & $0$\\ \hline
    \end{tabular}
\end{table}

\begin{figure}
    \centering
    \includegraphics[width=\linewidth]{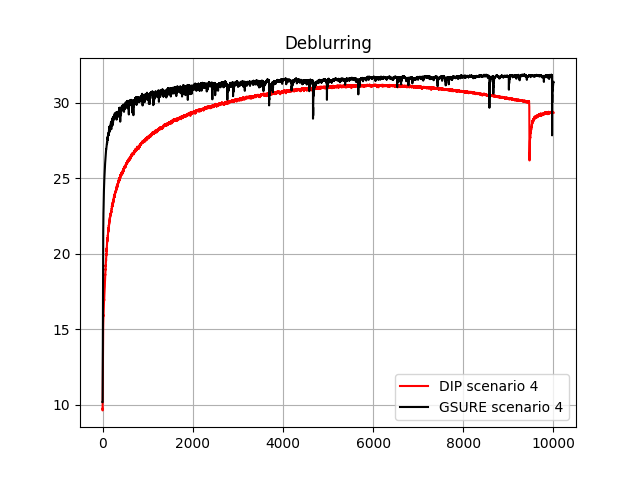}
    \caption{Deblurring PSNR results of scenario 4 averaged on Set5.}
    \label{fig:plot_deblur}
\end{figure}

Under this setup, we run GSURE for 4000 iterations and take the reconstruction that minimizes $\ell_{\scriptsize{GSURE}}(\Tilde{\x})$. Practically, GSURE does not need any early stopping criterion, since the divergence term in Equation~\eqref{eq:gsure} prevents the network from fitting the noise even when $\sigma$ is high. 
Therefore, the objective in Equation~\eqref{eq:gsure_cost_cnn} can be conveniently minimized until convergence, in contrast to DIP, where careful early stopping criteria need to be examined in order to achieve optimal results and prevent it from fitting the noise (as can be seen in Figure \ref{fig:plot_deblur}).

The average PSNR results of our method compared to DIP
are presented in the left columns of Table~\ref{table:deblurResults}. 
The superiority of our method over DIP is clear.

\begin{figure}[ht]
    \centering
    \begin{subfigure}{0.45\linewidth}
        \centering
        \includegraphics[trim={250pt, 0pt, 0pt, 250pt} ,clip,width=100pt]{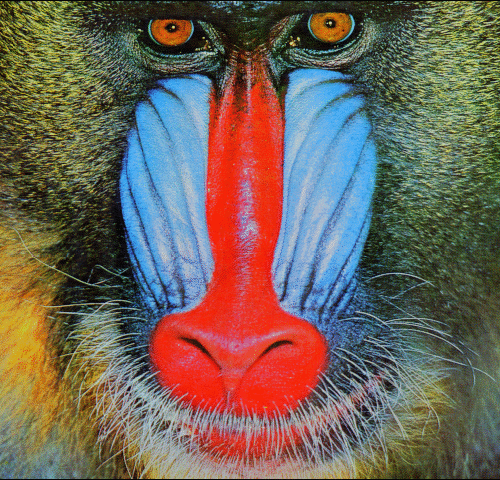}
        \caption{Original image (cropped)}
    \end{subfigure}
    \begin{subfigure}{0.45\linewidth}
        \centering
        \includegraphics[trim={250pt, 0pt, 0pt, 250pt}, clip,width=100pt]{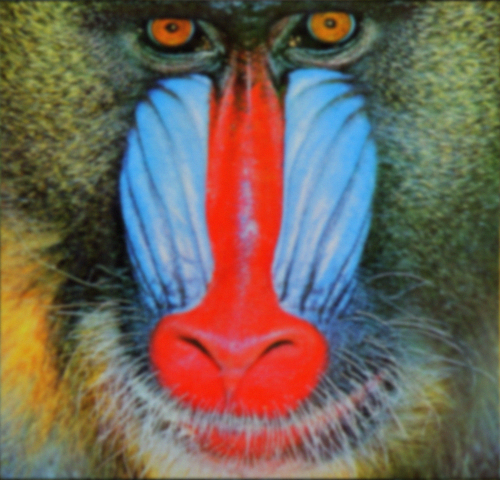}
        \caption{Blurred}
    \end{subfigure}\\
    \begin{subfigure}{0.45\linewidth}
        \centering
        \includegraphics[trim={250pt, 0pt, 0pt, 250pt} ,clip,width=100pt]{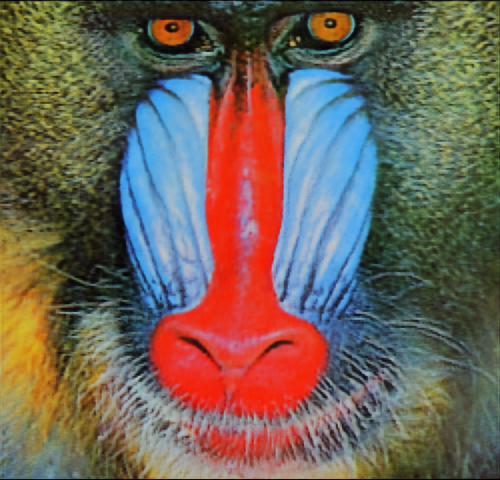}
        \caption{DIP}   
    \end{subfigure}
    \begin{subfigure}{0.45\linewidth}
        \centering
        \includegraphics[trim={250pt, 0pt, 0pt, 250pt} ,clip,width=100pt]{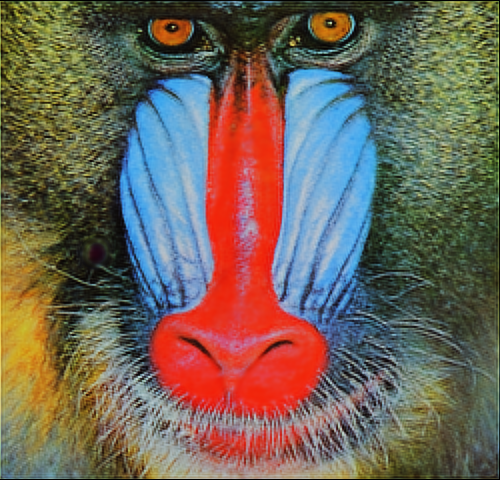}
        \caption{P-GSURE}       
    \end{subfigure}\\
    \begin{subfigure}{0.45\linewidth}
        \centering
        \includegraphics[trim={250pt, 0pt, 0pt, 250pt} ,clip,width=100pt]{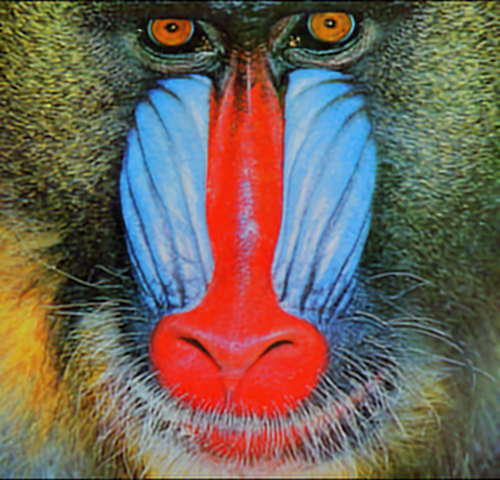}
        \caption{P\&P ADMM}   
    \end{subfigure}
    \begin{subfigure}{0.45\linewidth}
        \centering
        \includegraphics[trim={250pt, 0pt, 0pt, 250pt} ,clip,width=100pt]{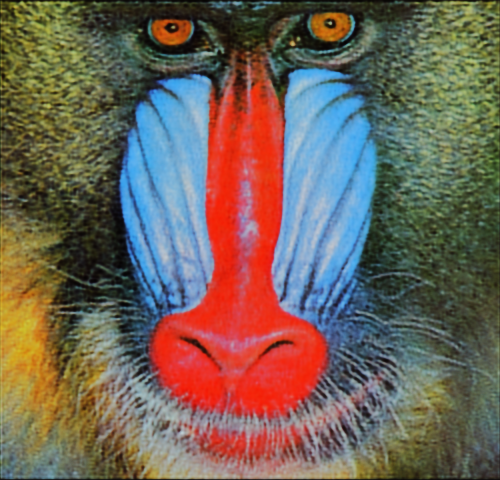}
        \caption{DIP-P\&P} 
    \end{subfigure}\\
    \begin{subfigure}{0.45\linewidth}
        \centering
        \includegraphics[trim={250pt, 0pt, 0pt, 250pt} ,clip,width=100pt]{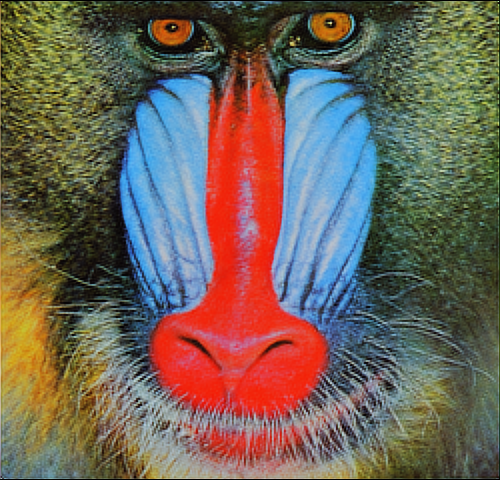}
        \caption{P-GSURE P\&P}  
    \end{subfigure}
    \caption{Deblurring of the baboon image from Set14, blurred using the model of scenario 5 from Table \ref{table:deblurScenarios}. (a) The original image cropped (b) The observed blurry and noisy image (c) Reconstruction using DIP (d) Reconstruction using P-GSURE (proposed method) (e) Reconstruction using P\&P-ADMM  method (using BM3D prior) (f) Reconstruction using DIP based P\&P-ADMM (g) Reconstruction using P-GSURE based P\&P-ADMM (proposed method).}
    \label{fig:comparison_deblur_scenario5}
\end{figure}

One may appreciate the visual quality of the GSURE based reconstruction by looking at Figure~\ref{fig:comparison_deblur_scenario2} and Figure~\ref{fig:comparison_deblur_scenario5}. The first is from scenario 2 and the second from scenario 5 in Table~\ref{table:deblurScenarios}.

We also examine the different methods within the ADMM-based P\&P scheme (presented with GSURE loss in Equation~\eqref{eq:gsure_admm}). In P\&P-GSURE, we set $\beta_i/\rho = 0.01$, where $\beta_i = \{0.75,0.75,4,1,2,1.5 \}$ and $i\in \{1,2,3,4,5,6\}$ indicates the scenario number. We initialize $\mathbf{v} = 0$ and $\z = \hat{\x}_{\scriptsize{ML}}$, and run 250 ADMM iterations, while performing the first step in \eqref{eq:gsure_admm} with 20 iterations of Adam with learning rate of $10^{-3}$. For P\&P-DIP we set $\beta/\rho = 1$, where $\beta=0.1$, initialize $\mathbf{v} = 0$ and $\z = \hat{\x}_{\scriptsize{ML}}$, and run 250 ADMM iterations, where in each iteration the first step in Equation~\eqref{eq:gsure_admm} is implemented using 20 iterations of Adam with learning rate set to $10^{-2}$. For the plain P\&P (without GSURE or DIP) we use $\beta_i = \{0.85,0.85,0.9,0.8,0.85,0.85\}$ and $\rho_i=\{ 2,1,3,1, 2, 2 \}/255$.
In all cases we implement the second step in Equation~\eqref{eq:gsure_admm} using the python version of the BM3D denoiser \cite{dabov2007image} with noise level $\sqrt{\beta/\rho}$. 

Because the projected GSURE measure is an estimate of the true projected MSE (up to a constant value), one could find the optimal hyper-parameters $\beta$ and $\rho$ that gives the lowest projected GSURE value. However, in the case of DIP, the true ground-truth image is required for tuning these parameters, which can be complicated.

The average PSNR results of combining the P\&P approach with GSURE and DIP are presented 
in the right columns of Table~\ref{table:deblurResults}.
It can be seen that incorporating GSURE and DIP with BM3D prior via the P\&P approach usually yields better results than their ``internal learning" application. Interestingly, P\&P-DIP is often no better than the plain P\&P scheme (which uses only the BM3D prior). Yet, the proposed P\&P-GSURE obtains the best results in most cases.

Finally, several visual examples of all the methods are displayed in Figure \ref{fig:comparison_deblur_scenario5}. It can be seen that our GSURE-based restoration yields sharper images with better textures.

\subsection{Super-Resolution}

In super-resolution $\H$ is defined as applying an anti-aliasing filter followed by sub-sampling by factor $\alpha > 1$. $\H$ is a circular convolution with filter $h$, therefore it can be applied efficiently using Fast Fourier Transform (FFT), then by sub-sampling the inverse FFT of the result we obtain the LR image. Accordingly, $\H^T$ in this case is defined as zero-padding of the input in between its samples, followed by applying a flipped version of $h$, which also can be applied using the FFT. Using the poly-phase decomposition (see, \eg  \cite{PnP_ADMM}) $\H\H^T$ can be achieved by multiplying the FFT of $h$ and its flipped version followed by sub-sampling the inverse FFT.

We test our approach on the test scenarios detailed in Table \ref{table:SRScenarios}. In these scenarios $\H$ is ill-posed, therefore when calculating $(\H\H^T)^\dagger$ we also zero all the eigenvalues smaller than $\xi$ (as in the deblurring case).

The average PSNR results of our method compared to DIP and ZSSR
are presented in the right columns of Table~\ref{table:SRResults}, and demonstrate the
superiority of our method.

Similar to the deblurring task, we examine the performance of DIP and GSURE within the ADMM-based  P\&P scheme. For P\&P-GSURE, we set $\beta = 100$ and $\beta/\rho = 10$, initialize $\mathbf{v} = 0$ and $\z = \hat{\x}_{\scriptsize{ML}}$, and run 50 ADMM iterations, while performing the first step in \eqref{eq:gsure_admm} with 100 iterations of Adam with learning rate of $10^{-3}$. In P\&P-DIP, we set $\beta/\rho = 1$, where $\beta = 0.1$ for scenarios \{1,3,4\} amd $\beta=1.5$ for scenarios \{2,5,6\}. We run 20 ADMM iterations, where in each iterations the first step in Equation~\eqref{eq:gsure_admm} is implemented using 250 iterations of Adam with learning rate set to $10^{-3}$. For the plain P\&P we use $\beta_i = \{0.85,0.9,0.85,0.85,0.9,0.9\}\times2$ and $\rho_i=\{2,3,2,2,3,3\}/255$. Similar to deblurring, we implement the second step in Equation~\eqref{eq:gsure_admm} using the python version of the BM3D denoiser \cite{dabov2007image} with noise level $\sqrt{\beta/\rho}$.

Similar to deblurring, note that the hyper-parameters $\beta$ and $\rho$ can be set by finding the minimum value of the projected GSURE, in contrast to the least-squares objective where one needs to tune them carefully in order to achieve optimal performance.

\begin{table*}
\small
\centering
    \caption{SR results (averaged PSNR) of the methods for the different scenarios} \label{table:SRResults}
    \begin{tabular}{ | c | c | c | c | c | c | c || c | c | c | c | c | c |}
    \hline
             & \multicolumn{3}{c|}{Set5} & \multicolumn{3}{c||}{Set14}                       & \multicolumn{3}{c|}{Set5} & \multicolumn{3}{c|}{Set14} \\ \hline
    Scenario     & DIP   & ZSSR             & GSURE         & DIP   &ZSSR   & GSURE         & P\&P & P\&P- & P\&P- & P\&P & P\&P- & P\&P- \\ 
    & & & & & & & & DIP & GSURE & &DIP &GSURE\\ \hline
        1        & 28.78 & 26.77            & \textbf{29.41}& 26.10 & 25.03 & \textbf{26.34}& 28.71 & 29.26 & \textbf{29.59} & 26.12 & 26.67 & \textbf{29.77} \\ \hline
        2        & 26.61 & 25.34            & \textbf{28.09}& 24.79 & 24.13 & \textbf{25.68}& 27.28 & 27.70 & \textbf{28.41} & 25.2  & 25.59 & \textbf{25.93} \\ \hline
        3        & 31.49 & \textbf{32.14}   & 32.04         & 28.38 & 26.07 & \textbf{29.05}& \textbf{32.83} & 31.92 & 32.79 & 29.57 & 29.14 & \textbf{29.58} \\ \hline
        4        & 29.49 & 27.33            & \textbf{29.48}& 26.14 & 25.23 & \textbf{26.65}& 29.82 & 29.95 & \textbf{29.92} & 26.90 & \textbf{27.19} & 27.00 \\ \hline
        5        & 29.67 & 29.71            & \textbf{31.09}& 27.08 & 25.21 & \textbf{28.32}& 30.43 & 31.15 & \textbf{31.46} & 27.90 & 28.47 & \textbf{28.77} \\ \hline
        6        & 27.60 & 25.97            & \textbf{28.79}& 25.41 & 24.45 & \textbf{26.20}& 27.92 & 28.51 & \textbf{28.94} & 25.63 & 26.30 & \textbf{26.46} \\ \hline
    \end{tabular} 
\end{table*}

\begin{table}
\small
\centering
    \caption{SR test scenarios} \label{table:SRScenarios}
    \begin{tabular}{ | c | c | c | c | c |}
    \hline
    Scenario     & $h(x_1, x_2)$& $\alpha$        & $\sigma^2$ & $\xi$ \\ \hline
        1        & Gaussian with $std = 1.6$    & 3             & 10  & $1\times 10^{-2}$\\ \hline
        2        & Gaussian with $std = 1.6$    & 3             & 49  & $1\times 10^{-2}$\\ \hline
        3        & Bicubic      & 2             & 10  & $1\times 10^{-2}$\\ \hline
        4        & Bicubic      & 3             & 10  & $1\times 10^{-2}$\\ \hline
        5        & Bicubic      & 2             & 49  & $1\times 10^{-2}$\\ \hline
        6        & Bicubic      & 3             & 49  & $1\times 10^{-2}$\\ \hline
    \end{tabular}
\end{table}

The average PSNR results of combining the P\&P approach with GSURE and DIP are presented in the left columns of Table~\ref{table:SRResults}.
Again, incorporating GSURE and DIP with BM3D prior via the P\&P approach usually yields better results than their ``internal learning" applications. Notice how again the proposed P\&P-GSURE obtains the best results in most scenarios.

Finally, several visual examples of all the methods are displayed in Figure \ref{fig:comparison_SR_scenario5}. It can be seen that our GSURE-based restoration yields sharper images with better textures.
\section{Conclusion}

In this work, we have presented a general framework for solving inverse problems using a neural network on a single image without the need for pre-training on a large dataset. Our approach relies on an unbiased estimator of the MSE, namely the generalized SURE that approximates the true error. This allows training a network directly on the input image without external data.

Our approach improves over the DIP strategy, which train networks based on the least-squares loss. Moreover, it has the advantage that it does not need to rely on early stopping and less sensitive to the used architecture. In the work, for the sake of comparison, we have used the same architecture proposed in the DIP work. Yet, when we changed the network model, our approach still achieved good results while the performance of DIP decreased more significantly.

While in this work we focus on using the GSURE formulation for white Gaussian noise, it might be possible in follow-up work to support a larger variety of noise types using the formulation in \cite{Eldar2009fga} that supports also colored noise and other types of noise. Moreover, while our work focused on deblurring and super-resolution, our proposed framework can be easily used also for other inverse problems in image processing such as joint demosaicing and denoising.

\subsubsection*{Acknowledgments}

The work is supported by the ERC-StG (No. 757497) grants.

\small

{\small
\bibliographystyle{ieee_fullname}
\bibliography{egbib}
}

\end{document}